\documentclass[runningheads]{llncs}
\usepackage{lineno}
\usepackage[T1]{fontenc}
\usepackage{xspace}
\usepackage{amsmath}

\newcommand{\eg}{e.g.\xspace}
\newcommand{\ie}{i.e.\xspace}
\usepackage{graphicx}
\usepackage{booktabs}
\usepackage{epstopdf}
\usepackage{float}
\usepackage{multirow}
\usepackage{chngcntr}
\usepackage{hyperref}
\usepackage{cleveref}
\begin{document}
\title{Assessing the Visual Enumeration Abilities of Specialized Counting Architectures and Vision-Language Models}
\titlerunning{Enumeration Abilities of Specialized Counting Architectures and VLMs}
\author{Kuinan Hou\inst{1} \and
Jing Mi\inst{1}\ \and
Marco Zorzi\inst{1}\ \and
Lamberto Ballan\inst{1}\ \and
Alberto Testolin\inst{1}} 
\authorrunning{Hou et al.}
%
\institute{University of Padova, 35131 Padova, IT}
\maketitle              
\begin{abstract}
Counting the number of items in a visual scene remains a fundamental yet challenging task in computer vision. Traditional approaches to solving this problem rely on domain-specific counting architectures, which are trained using datasets annotated with a predefined set of object categories. However, recent progress in creating large-scale multimodal vision-language models (VLMs) suggests that these domain-general architectures may offer a flexible alternative for open-set object counting. In this study, we therefore systematically compare the performance of state-of-the-art specialized counting architectures against VLMs on two popular counting datasets, as well as on a novel benchmark specifically created to have a finer-grained control over the visual properties of test images. Our findings show that most VLMs can approximately enumerate the number of items in a visual scene, matching or even surpassing the performance of specialized computer vision architectures. Notably, enumeration accuracy significantly improves when VLMs are prompted to generate intermediate representations (i.e., locations and verbal labels) of each object to be counted. Nevertheless, none of the models can reliably count the number of objects in complex visual scenes, showing that further research is still needed to create AI systems that can reliably deploy counting procedures in realistic environments.

\end{abstract}

\section{Introduction}
\label{sec:intro}

Object counting—the task of identifying and enumerating objects of interest in a visual scene—is a fundamental challenge in computer vision, with applications ranging from ecological monitoring to retail inventory management.
Notably, many animal species can approximately estimate the number of objects in visual scenes, with estimation errors scaling proportionally with number according to Weber's law \cite{feigenson2004core,testolin2021estimates}, but only educated humans can deploy systematic counting procedures to derive the cardinality of sets in an exact manner \cite{gallistel1992preverbal}.

Despite its apparent simplicity, visual enumeration remains challenging even for the most advanced AI systems \cite{testolin2025visual}.
Furthermore, although modern counting architectures can work in constrained settings \cite{dai2023cross,gomaa2022faster}, they struggle in open-world scenarios where objects span a wide range of diverse categories, appear in cluttered environments, or belong to novel classes that were not present during model training.
This happens because most counting approaches rely on category-specific object detection and localization techniques \cite{trott2017interpretable,zhang2018learning,babu2022completely,kilic2023accurate}, limiting their generalization capabilities.

Recently proposed text-prompting methods \cite{huang2024point,shi2024training,jiang2023clip} partially mitigate this issue by allowing users to specify target categories via natural language. These approaches typically rely on CLIP-like architectures, where a vision encoder processes the input image, and a language encoder parses the textual prompt, thus exploiting the vast amount of knowledge embedded in large-scale foundation models \cite{bommasani2021opportunities}.
In this work, we further explore the potential of these domain-general models by investigating whether recent multimodal vision-language models (VLMs) can, in fact, directly tackle open-set object counting when properly prompted. To this aim, we consider the most advanced VLMs currently available, namely Claude 4.5 Sonnet, Gemini 2.5 Pro and GPT-5, as well as the open-source Qwen3-VL model, and compare their enumeration performance against state-of-the-art computer vision architectures specifically developed for object counting.

We evaluate all models on three distinct datasets, each representing a different level of visual and contextual complexity. This multi-dataset evaluation enables a comprehensive assessment of model performance across a range of realistic and controlled scenarios, at the same time highlighting potential limitations of current reference benchmarks.
We also show that cognitively-inspired prompt engineering can drastically improve the enumeration skills of some VLMs: in particular, Gemini achieves state-of-the-art performance across all benchmarks when asked to explicitly locate and label each object to be counted.
Overall, by guiding VLMs to reason about how to count we thus demonstrate that multimodal AI systems can deploy enumeration strategies that allow them to surpass the performance of traditional category-specific counting pipelines.

\section{Methodology}
\subsection{Evaluation Datasets}

We address the problem of category-agnostic object counting in images depicting a wide variety of visual environments, focusing on numerosities $\leq 40$ to align with the limits of human numerosity estimation \cite{anobile2014separate} and to guarantee a fair comparison even when models are trained on images with a limited number of objects. Indeed, the distribution of countable objects in large-scale image datasets used to train foundation models typically shows a distinctive power-law trend \cite{hou2025estimating}, implying that small numbers are overrepresented compared to larger numbers. 

\subsubsection{FSC-147 and FSCD-LVIS}

We employ FSC-147 \cite{ranjan2021learning} and FSCD-LVIS \cite{nguyen2022few} as reference benchmarks for object counting in naturalistic settings. FSC-147 is a widely used few-shot counting dataset that contains diverse object categories and varying object densities, annotated with point-level supervision. FSCD-LVIS extends this setup to a long-tailed distribution of object classes derived from the LVIS ontology, thus posing additional challenges related to class imbalance and open-set generalization. These datasets are intended to test the ability to handle real-world clutter, scale variation, and semantic diversity. However, as we will discuss below, it turns out that FSCD-LVIS often contains extremely challenging images, which may not be adequate for objectively assessing the performance of counting models (see examples in Fig.~\ref*{fig:fsc_example} and Fig.~\ref*{fig:FSCD-LVIS Difficult Samples} in the Supplementary).

\begin{figure}[]
    \centering
    \includegraphics[width=0.96 \linewidth]{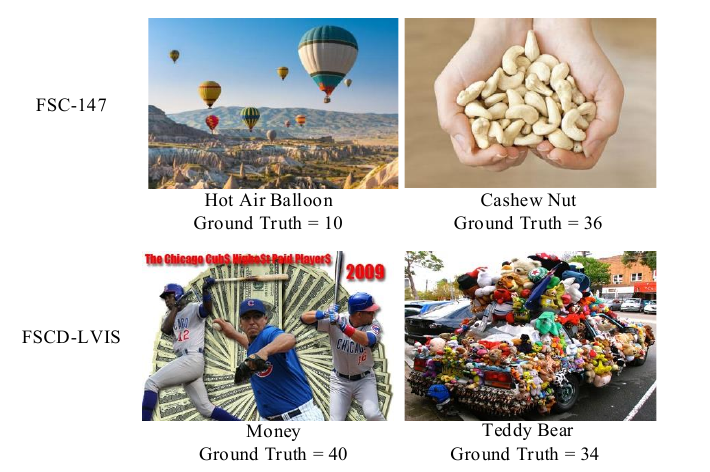}
    \caption{Challenging examples from the FSC-147 and FSCD-LIVS datasets.}
    \label{fig:fsc_example}
\end{figure}

\subsubsection{SolidCount}
\begin{figure}[ht]
    \centering
    \includegraphics[width=1\linewidth]{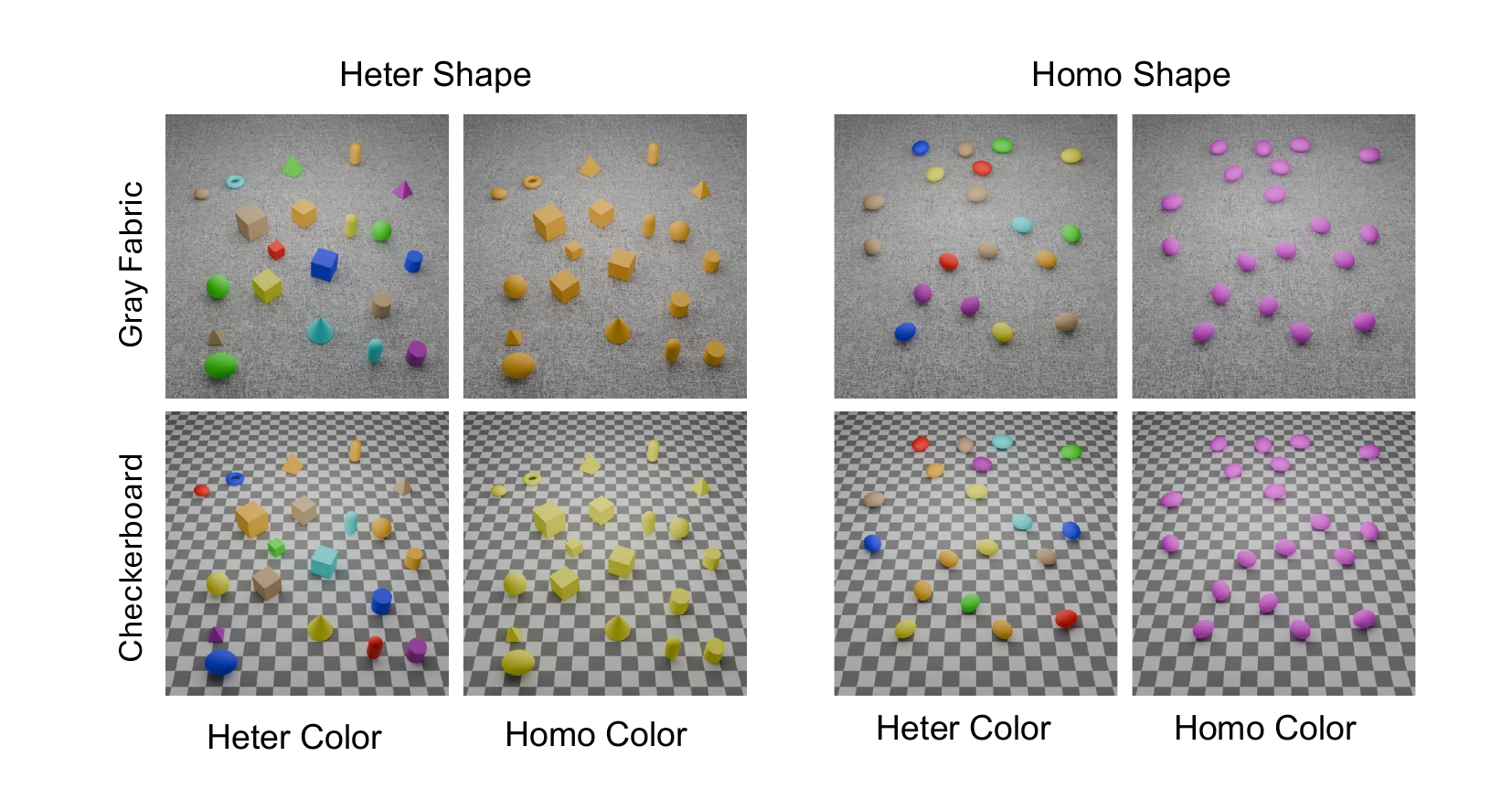}
    \caption{Examples of synthetic images from the SolidCount dataset. Rows display the two different backgrounds; odd/even columns represent heterogeneous/homogeneous colors, while left/right panels show images with heterogeneous/homogeneous shapes.}
    \label{fig:sc_illu}
\end{figure}

To assess visual enumeration abilities under more precise, tunable conditions, we introduce SolidCount, a synthetic benchmark generated via a custom Blender pipeline. Images contain 8--40 objects drawn from eight primitive shapes and eight saturated colors with calibrated lighting. Objects are randomized in location, rotation, and scale (projected to 3--15\% of image height), subject to strict constraints to minimize occlusion: minimum world-space separation of 0.15 units, maximum pixel-wise overlap of 1\%, and full visibility within the central 90\% of the view\footnote{We make SolidCound freely available to download \href{https://drive.google.com/drive/folders/1i4lWBaX9IVBBUtvV1L-qw5ItASMeqsJn?usp=sharing}{here}.}.
This allowed us to systematically investigate how visual complexity affects counting performance by isolating a variety of features: background type (gray fabric vs. high-contrast checkerboard) and heterogeneous vs. homogeneous shapes and colors (see examples in Figure~\ref{fig:sc_illu}).

\subsection{Problem Formulation and Evaluation Metrics}
Given an input image $I$, the goal is to estimate the count $C$ of objects belonging to a target category $T$. To obtain the estimated $\hat{C}_i$, we use different strategies depending on the nature of the model considered. We evaluate counting performance using two complementary metrics: mean absolute error (MAE) and accuracy. MAE measures the average absolute deviation between the predicted count and the ground-truth count, computed as $\text{MAE} = \frac{1}{n}\sum_{i=1}^{n}|C_i - \hat{C}_i|$ for $n$ test samples. Accuracy reflects the proportion of predictions whose estimated count exactly matches the ground-truth value, providing a direct assessment of correct predictions. 

\subsection{Models Considered}  

We evaluate two groups of models: 1) models specifically designed for object counting and 2) domain-general vision-language models (VLMs) that can exhibit a variety of visual reasoning skills.  

\subsubsection{Counting Models}  
We consider three state-of-the-art counting architectures: Point, Segment, and Count (PseCo) \cite{huang2024point} is a detection-based architecture that utilizes point-level supervision and 
segmentation cues to localize and enumerate objects; Training-Free Object Counting (TFOC) \cite{shi2024training} is also a detection-based architecture that leverages pre-trained feature extractors, and T2ICount \cite{qian2025t2icount} is a recently introduced regression-based architecture.

\subsubsection{Multimodal Vision-Language Models (VLMs)}  
We consider four large-scale multimodal models representing the state-of-the-art in AI research: the multimodal Claude 4.5 Sonnet model [claude-sonnet-4-5@20250929] recently developed by Anthropic, the multimodal Gemini 2.5 Pro model [gemini-2.5-pro] developed by Google, and the multimodal GPT-5 model [gpt-5-2025-08-07] developed by OpenAI \cite{openai2024gpt4ocard}. The open-source multimodal Qwen3-VL [qwen3-vl-235b-a22b-Instruct] was also evaluated, but give its overall low performance it was omitted in some analyses.


\subsection{Prompting Methods}

For counting models, prompts consisted solely of the category name of the target object. These names were sourced from the annotation files of the FSC-147 and FSCD-LVIS datasets. For the SolidCount dataset, we use ``geometric shapes" as category name in consistent with the VLMs.
For VLMs, we employed a structured system message—following guidelines from Gemini’s bounding box detection documentation\footnote{\url{https://cloud.google.com/vertex-ai/generative-ai/docs/bounding-box-detection}}—to enforce consistent output formatting. The user message always specified the category name and asked how many instances of the target object appeared in the image (details in Section~\ref*{sec:prompts} of the Supplementary Material). We evaluated three distinct VLM prompting strategies, differing in the explicitness of the counting procedure:

\begin{itemize}
    \item \textbf{Naive count}: The model is directly asked to return the total number of target objects, without any intermediate representation.
    \item \textbf{Label and count}: The model explicitly enumerates all target objects by generating a unique label for each and returning both the list of labels and the total count.
    \item \textbf{Point, label and count}: The model first localizes each target object by generating bounding boxes, then assigns a unique label to each, and finally returns the total count by summing up the number of boxes.
\end{itemize}

\noindent
This prompt hierarchy is inspired by increasingly more sophisticated counting strategies used by humans: naive counting resembles rapid, parallel estimation under time constraints \cite{dehaene2011number}, while the more structured approaches emulate serial, symbolic counting used when precise enumeration is required \cite{gallistel1992preverbal,carey2019ontogenetic}. By comparing these methods, we assess whether VLMs achieve better counting accuracy when guided to explicitly individuate and/or localize objects.

\section{Results}
\label{sec:results}

\begin{figure}[t]
    \centering
    \includegraphics[width=0.99\linewidth]{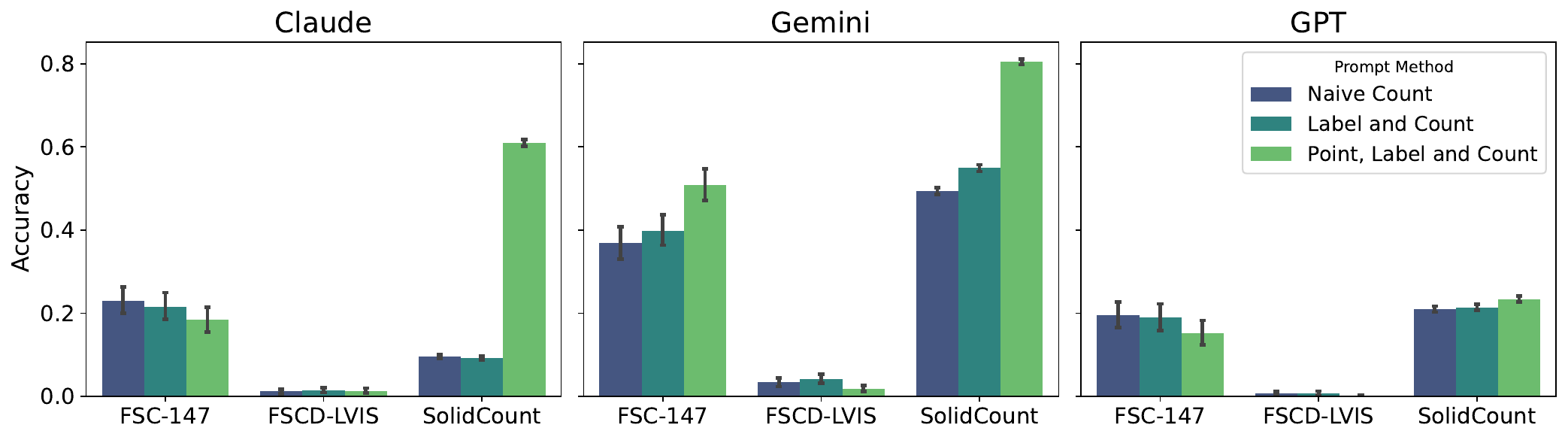}
    \caption{Bar chart displaying the accuracy of three prompting strategies across all datasets. The height of each bar represents the mean accuracy, and the error bars indicate the 95\% confidence interval of the mean.}
    \label{fig:acc_prompt_method}
\end{figure}

Gemini was by far the top-performing VLMs across all datasets. Furthermore, as shown in Fig. \ref{fig:acc_prompt_method}, in most of the cases the "Point, label and count" prompt allowed to achieve dramatic improvements in counting accuracy, demonstrating that guiding the models to first locate and identify each object yields a more robust enumeration behavior (the Qwen model was not able to follow the structured prompt instructions and was therefore not considered in this analysis).
On SolidCount, the "Point, label, and count" strategy yielded significant improvements for Claude and Gemini, but not for GPT. Claude’s effect size was large ($\epsilon^2 = 0.291$), while Gemini’s was medium ($\epsilon^2 = 0.109$). GPT showed no significant difference across prompting strategies.
On the real-image datasets, results were more mixed. On FSC-147, only Gemini showed a significant improvement with the more structured prompt, with a small effect size ($\epsilon^2 = 0.014$). Claude and GPT showed no significant differences. On FSCD-LVIS, none of the models exhibited significant differences across prompting strategies, consistent with the uniformly low performance on this challenging dataset. The trends for MAE largely mirrored the phenomenon observed for accuracy (see Figure~\ref{fig:mae_prompt_method}). 

Table \ref{tab_best_mae_acc_real} summarizes the performance of all models across datasets (for VLMs we considered the best performing prompting strategy). On FSC-147, VLMs demonstrate remarkable zero-shot capability, with Gemini achieving an MAE of 2.43, surpassing all specialized counting architectures. This advantage is even more pronounced on SolidCount, where all specialized architectures perform poorly, despite the simplicity of the synthetic visual scenes. The enumeration performance of all models collapses on the FSCD-LVIS dataset, where ambiguous annotations and class imbalance present significant challenges. Upon further inspection of the images and annotations, we found that many ground truth annotations are questionable due to factors such as objects being cropped at image boundaries, low object salience, and the presence of ambiguous or even non-countable target categories (see Section~\ref*{sec:fscd_challenge} in the Supplementary Material), suggesting that this might not be an appropriate benchmark to evaluate visual enumeration.
Notably, all models obtain a remarkably low accuracy across all benchmarks, suggesting that in most cases they cannot systematically count all objects in the visual scene, but rather just provide an approximate guess of the numerosity.

\begin{table}
\caption{Mean Absolute Error and accuracy for each model on the three benchmarks. The most accurate model for each dataset was always Gemini.}
\label{tab_best_mae_acc_real}
\centering
\begin{tabular}{|l|cc|cc|cc|cc|cc|cc|cc|}
\hline
Dataset & \multicolumn{2}{c|}{Claude} & \multicolumn{2}{c|}{Gemini} & \multicolumn{2}{c|}{GPT-5} & \multicolumn{2}{c|}{Qwen} & \multicolumn{2}{c|}{PseCo} & \multicolumn{2}{c|}{T2ICount} & \multicolumn{2}{c|}{TFOC} \\
\cline{2-15}
~ & MAE & Acc. & MAE & Acc. & MAE & Acc. & MAE & Acc. & MAE & Acc. & MAE & Acc. & MAE & Acc. \\
\hline
FSC-147
& 3.39 & 0.21
& \textbf{2.43} & \textbf{0.51}
& 3.46 & 0.20
& 3.66 & 0.23
& 31.77 & 0.01
& 3.66 & 0.38
& 7.78 & 0.14 \\
\hline
FSCD-LVIS
& 13.72 & 0.01
& \textbf{14.66} & \textbf{0.04}
& 15.86 & 0.01
& 18.42 & 0.00
& 108.46 & 0.003
& 24.05 & 0.02
& 19.53 & 0.01 \\
\hline
SolidCount
& 0.64 & 0.61
& \textbf{0.41} & \textbf{0.80}
& 2.07 & 0.21
& 1.67 & 0.30
& 253.85 & 0.06
& 67.53 & 0.03
& 192.05 & 0.03 \\
\hline
\end{tabular}
\end{table}

\begin{figure}[H]
    \centering
    \includegraphics[width=0.8\linewidth]{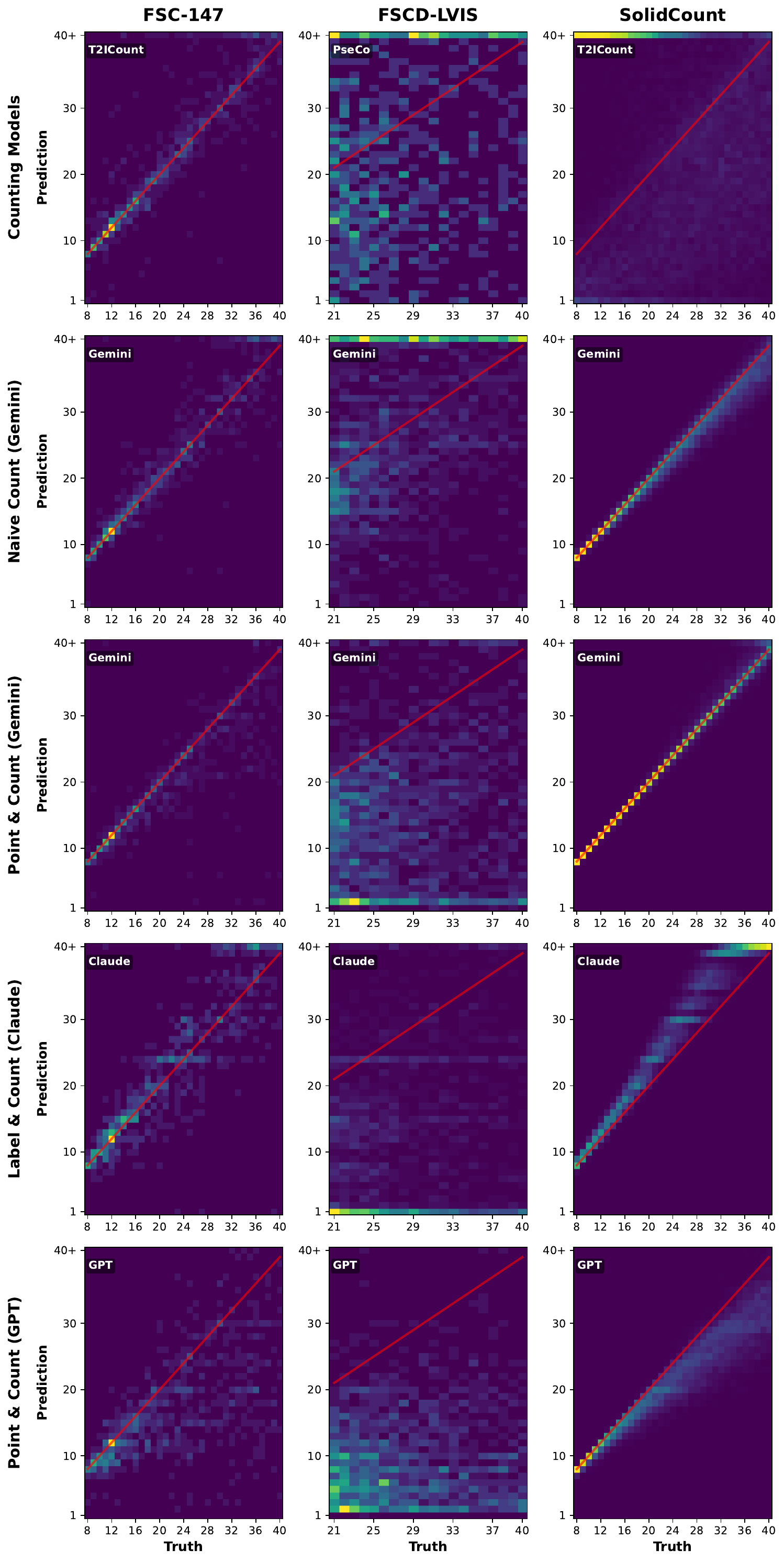}
    \caption{Confusion matrices across all datasets for the best counting specific models and representative VLMs. The x-axis represents the ground truth, while the y-axis denotes the predictions, with the final row ("40+") aggregating all predictions where $n > 40$. The red line indicates the perfect responses.}
    \label{fig:cm_best}
\end{figure}

Figure \ref{fig:cm_best} provides a visual analysis of the error distributions for the best counting specific architecture and VLM on each dataset. The confusion matrices for FSC-147 and SolidCount corroborate the quantitative evaluation metrics: the best VLM (Gemini) exhibits tight convergence along the diagonal (especially when using the "point, locate and count" prompt), while specialized models dramatically fail even on synthetic visual scenes. Claude and GPT showed opposite trends on SolidCount, with Claude underestimating and GPT overestimating images featuring many objects. On FSCD-LVIS, all models exhibit significant dispersion—a widening ``cloud'' around the diagonal rather than a tight band. This pervasive variance suggests that errors stem less from model incapacity and more from the inherent noise and ambiguity of this dataset. Confusion matrices for all models are reported in the Supplementary Materials.

We quantitatively analyzed how model performance varied as a function of the number of objects to count. On real-image benchmarks (FSC-147 and FSCD-LVIS), specialized counting models generally exhibited lower sensitivity to increasing object counts than VLMs. Specifically, counting models showed MAE regression slopes ranging from $0.10$--$0.23$ on FSC-147 and $0.41$--$0.91$ on FSCD-LVIS, while VLMs degraded more rapidly with increasing numerosity, with slopes of $0.24$--$0.32$ on FSC-147 and $0.72$--$0.99$ on FSCD-LVIS.
On SolidCount, however, this trend reversed. VLMs demonstrated strong robustness to increasing object counts, with minimal MAE slopes ($0.06$--$0.20$), indicating relatively stable error as numerosity grows. Conversely, counting models exhibit large negative slopes (ranging from $-7.36$ to $-3.63$). This inverse relationship suggests that, while VLMs maintain consistent performance, counting models suffer from severe scaling artifacts on synthetic images, driven by systematic over-prediction in certain count ranges rather than genuine adaptability.

\subsection{Factors affecting Counting Performance}
\begin{figure}
    \centering
    \includegraphics[width=0.99\linewidth]{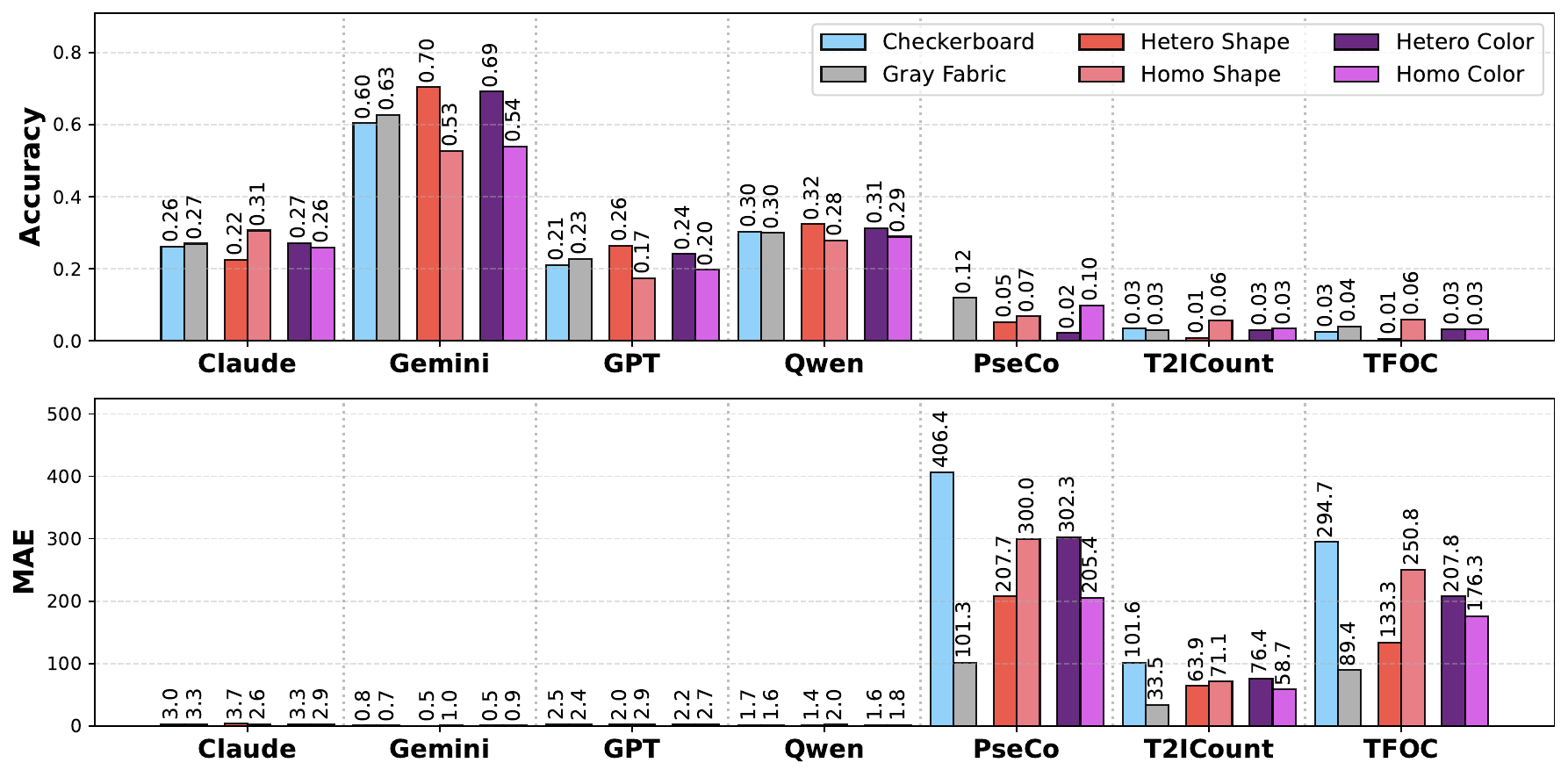}
    \caption{The figure presents Accuracy (top) and MAE (bottom) for seven models. For each model, results are organized into three paired comparisons: Background (Checkerboard vs. Gray Fabric), Shape (Heterogeneous vs. Homogeneous), and Color (Heterogeneous vs. Homogeneous).For VLMs that have more than one prompt methods, we show the average of all prompt methods.}
    \label{fig:counting_factors}
\end{figure}

While real-world data provides a faithful baseline, the SolidCount dataset allows us to investigate which visual features mostly impact counting performance. Figure~\ref{fig:counting_factors} shows how accuracy and MAE change as a function of specific manipulations in background, shapes, and colors variability, which we separately analyse and discuss below.

\subsubsection{Background}
\begin{figure}
    \centering
    \includegraphics[width=0.80\linewidth]{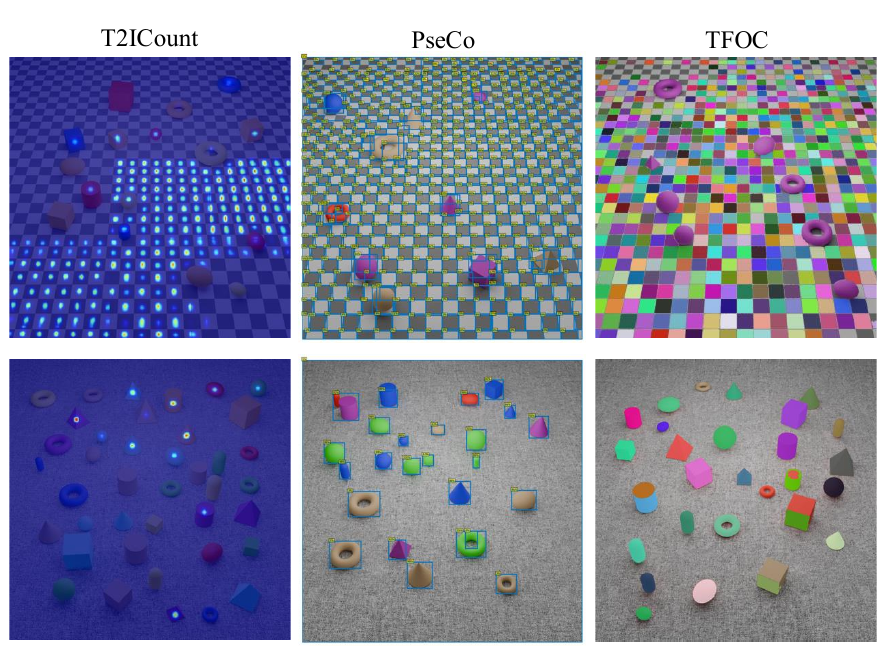}
    \caption{Visualization of the failure of counting specific models in detecting relevant objects due to the confounding background (upper panels), which causes the models to erroneously count also checkerboards.}
    \label{fig:error_background}
\end{figure}

To quantify the influence of environmental clutter, we compared counting performance under a neutral background (uniform gray fabric) and a challenging background (high-frequency checkerboard), analyzing effects on a per-model basis.
Among VLMs, background clutter induced only modest and model-dependent changes in performance. Gemini, GPT, and Qwen exhibited small but statistically significant increases in MAE under the checkerboard condition (Gemini: 0.71 $\to$ 0.76, $p=0.027$, $d=0.022$; GPT: 2.36 $\to$ 2.51, $p<0.001$, $d=0.061$; Qwen: 1.63 $\to$ 1.72, $p=0.004$, $d=0.050$), accompanied by minor reductions in accuracy (odds ratios ranging from 0.906 to 0.907). Claude followed a slightly different pattern, showing a small reduction in MAE under clutter (3.27 $\to$ 2.99; $p<0.001$, $d=-0.083$), with a marginal and statistically weak decrease in accuracy ($\text{OR}=0.957$, $p=0.054$). Overall, these effects indicate that while clutter mildly perturbs VLM performance, error magnitudes remain stable and do not exhibit catastrophic failure.
In contrast, specialized counting models were severely impacted by background clutter. They all showed dramatic MAE inflation under checkerboard backgrounds (e.g., PseCo: 101.32 $\to$ 406.37, $p<0.001$, $d=1.893$; TFOC: 89.35 $\to$ 294.74, $p<0.001$, $d=1.030$), coupled with substantial drops in accuracy. Notably, PseCo\ collapsed entirely under clutter, with accuracy falling from 11.9\% to 0.0\% ($\text{OR}=0.001$). T2ICount\ similarly exhibited a MAE explosion (33.47 $\to$ 101.60; $p<0.001$, $d=0.574$), though accuracy changes were less pronounced as the performance was already near the floor. These consistent trends suggest that high-frequency textures trigger massive false positive counts in proposal-based pipelines, as also shown in the examples in Figure \ref{fig:error_background}.

\subsubsection{Heterogeneous Shapes}
To evaluate the impact of shape heterogeneity, we compared the counting performance when images only contained heterogeneous shapes (i.e., at least 7 or 8 different solids) against the condition with homogeneous shapes (for examples, see left vs. right panels in Figure~\ref{fig:sc_illu}).
An interesting ``preference for shape diversity'' was observed in almost all VLMs, while traditional counting models performed better when images contained uniform shapes. In particular, Gemini's MAE decreased from 0.99 (homogeneous) to 0.48 (heterogeneous; $p < 0.001$, $d=-0.270$), with accuracy increasing from 52.7\% to 70.4\% ($\text{OR}=2.141$). This suggests that VLMs might tend to confuse objects with the same shape (e.g., by double counting the same object or missing some).
In contrast, counting models generally floundered when faced with shape diversity. PseCo\ exhibited lower accuracy under heterogeneous shapes (5.1\% vs.\ 6.8\%; $p < 0.001$, $\text{OR}=0.741$), despite a reduction in MAE (207.70 vs.\ 300.00; $d=-0.425$). T2ICount accuracy dropped sharply from 5.6\% (Homogeneous) to 0.9\% (Heterogeneous; $p < 0.001$, $\text{OR}=0.148$), and TFOC\ followed the same pattern (Accuracy $5.9\% \to 0.6\%$; $p < 0.001$, $\text{OR}=0.090$), confirming that for specialized counters, shape heterogeneity impairs the proposal generation process.

\subsubsection{Heterogeneous Colors}
Finally, we examined the impact of color heterogeneity by comparing performance on the subset of images with heterogeneous versus homogeneous colors, while keeping geometric configurations and background settings constant.
The divergence between VLMs and counting models was most pronounced here. Similarly to shape diversity, VLMs demonstrated better performance for images with color diversity. Gemini's performance significantly improved in the heterogeneous color condition (MAE $0.55$ vs.\ $0.92$; $p < 0.001$, $d=-0.198$), with significantly higher odds of success ($\text{OR}=1.916$). GPT similarly achieved lower MAE (2.19 vs.\ 2.68; $p < 0.001$, $d=-0.201$) and higher accuracy (24.1\% vs.\ 19.7\%; $\text{OR}=1.292$) with diverse colors. Qwen also benefited from heterogeneity (MAE $1.56$ vs.\ $1.78$; $p < 0.001$, $d=-0.121$). This suggests that color variation provides critical perceptual information that helps VLMs segment and individuate objects. This is in contrast with recent human studies showing that an increase in the heterogeneity of color in the visual scene causes an underestimation of numerosity \cite{li2025influence,qu2022increasing}.
Conversely, counting models performed better under color homogeneity. PseCo\ performed significantly worse with heterogeneous colors (MAE 302.30 vs.\ 205.39; $p < 0.001$, $d=0.448$), with accuracy dropping sharply (2.2\% vs.\ 9.8\%; $\text{OR}=0.203$). T2ICount\ similarly saw reduced accuracy in the heterogeneous condition (3.1\% vs.\ 3.4\%; $\text{OR}=0.907$), while TFOC\ showed higher error under color heterogeneity (MAE 207.79 vs.\ 176.30; $p < 0.001$, $d=0.141$). These specialized architectures rely on consistent visual features to cluster object instances, a mechanism that seems impaired by color heterogeneity.

\section{Discussion}
\label{sec:discussion}

The results presented in this work highlight that despite notable architectural advances, object counting remains a challenging task even for the most capable AI systems. However, some state-of-the-art models, such as Claude 4.5 Sonnet and Gemini 2.5 Pro, already surpass counting-specific architectures on most benchmark datasets, especially when prompted with more structured instructions.
These findings suggest that large-scale multimodal systems may soon replace specialized architectures tailored for counting, which rely on dense regression \cite{xie2018microscopy} or detection pipelines \cite{huang2024point} to achieve high localization accuracy but suffer from limited generalization beyond predefined object categories. Indeed, the performance of counting-specific models tends to degrade in scenes with high clutter, occlusion, or category ambiguity.

The factorial analysis we performed on our newly introduced dataset (SolidCount) reveals a fundamental mechanistic divergence between counting-specific models and VLMs. The catastrophic failure of counting models on checkerboard backgrounds is driven by the over-generation of object proposals in background regions: although all three counting models utilize advanced vision transformer backbones or SAM-based decoders~\cite{kirillov2023segment}, they remain heavily reliant on bottom-up ``objectness'' cues to generate candidate masks. The repetitive, high-frequency structure of the checkerboard acts as a super-stimulus for these object detectors, leading to massive false-positives. This vulnerability to background clutter is a known limitation in regression- and detection-based counting paradigms, where models frequently fail to suppress predictions in texture-rich regions \cite{modolo2020understandingimpactmistakesbackground}. In contrast, VLMs exhibit robust semantic grounding. The textual prompt acts as a stronger top-down semantic filter, effectively suppressing visually salient but semantically irrelevant regions. Furthermore, the VLM preference for heterogeneous shapes and colors suggests that visual diversity aids these models in individuation—providing distinct semantic anchors that prevent the token merging (or "attention collapse") often observed when processing uniform, repetitive sets. Recent work on text-only sequential enumeration tasks also observed this preference for heterogeneity in large language models, where non-uniform strings are counted more accurately \cite{hou2025sequential}.

Furthermore, our work shows that incorporating explicit detection supervision during training (\eg, Gemini) and using more structured prompting (\eg, Gemini and Claude) can dramatically enhance visual reasoning capabilities. In particular, the ``point, label and count'' and ``label and count'' approaches have an intrinsic serial nature, which implies that the image can be revisited multiple times during token generation. Though the exact mechanisms employed by the different VLMs remain to be investigated, the procedure of serially tagging each object using a verbal label is indeed at the heart of human counting \cite{gelman2009child}. 

Future work should explore how generation parameters (\eg, temperature, top-k, top-p) affect the precision and reproducibility of coordinate outputs, and investigate the neuronal differences in activation-based localization and token-based generation.

\section{Conclusion}

A notable methodological contribution of the present work has been the introduction of SolidCount, a synthetic benchmark that addresses significant limitations in existing counting datasets and allows to systematically assess the impact of visual features (such as background clutter, shape and color diversity) on the enumeration performance. Our investigation also revealed that widely used benchmarks such as FSCD-LVIS contain noisy images with questionable ground truth annotations, and might therefore not be suited to precisely characterize enumeration capabilities.

As for the architecture of counting models, transformer-based encoders underpin the success of both counting-specific architectures and multimodal VLMs, and models leveraging segmentation \cite{kirillov2023segment} and CLIP \cite{radford2021learning} modules demonstrate that shared encoder blocks can help extract generalizable features across diverse object categories and visual scenes. This aligns with broader trends in multimodal AI: architectures like Janus Pro \cite{chen2025janus} show that task-specific decoders paired with unified encoders enable versatility in handling distinct visual tasks (\ie image understanding vs. image generation), suggesting that advancing transformer-based encoders (rather than developing isolated task-specific architectures) may drive further progress toward human-level visual enumeration capabilities.

Finally, a novel finding of this work is that VLMs can exploit task decomposition and the generation of intermediate verbal representations to significantly improve counting accuracy. Nevertheless, in agreement with other recent findings \cite{testolin2025visual}, we conclude that at the moment visual counting still remains a key and challenging benchmark for evaluating AI’s visual reasoning capabilities.

\subsubsection{Acknowledgements} This project was partially supported by the Italian Ministry of Education and Research [PRIN Grant 2022EBC78W] and by the European Union - NextGenerationEU as part of the National Recovery and Resilience Plan (PNRR) [Project GROUNDEEP, FAIR Missione 4, Componente 2, CUP J93C24000320007]. K.H. acknowledges the support of the China Scholarship Council (ID: 202307820031).

%
%
%
\bibliographystyle{splncs04}
\bibliography{main}

\clearpage
\setcounter{page}{1}
\setcounter{section}{0}
\setcounter{figure}{0}
\setcounter{table}{0}

\section*{Supplementary Material}

\setcounter{figure}{0}
\makeatletter 
\renewcommand{\thefigure}{S\@arabic\c@figure}
\makeatother

\section{Data Availability}\label{sec:data_avail}
We provide all the inference code and results in the zip file during the revision process. Once published, we will make all data and code available online.

\section{Prompts}\label{sec:prompts}

\subsection{Prompts for VLMs}
Here we present the final prompts used for prompting the VLMs.
System prompts for estimation approach:
\begin{quote}
``You are attending a numerical perception test, perform your best. Only respond with a single number for your count result and nothing else.''
\end{quote}

System prompts for localizing approach:
\begin{quote}
``Return bounding boxes as an array with labels.  
Never return masks. The bounding boxes should be in \texttt{Ymin, Xmin, Ymax, Xmax} format, with coordinates normalized from 0 to 1000.  
If an object appears multiple times, assign each instance a unique label based on its distinct characteristics (e.g., color, size, position, etc.).''
\end{quote}

While for the labeling approach, the system message was:
\begin{quote}
``You are attending a counting ability test, perform your best counting ability. You are going to count certain objects given an image. Count from left to right, top to bottom. Assign each object a unique label. Return the labels as an array of strings and an integer indicating how many objects there are in total''
\end{quote}

\noindent
Then, for each image, we paired it with a user message specifying the object category:

\begin{quote}
``Output the positions of all the \texttt{[category name]} in the image.''
\end{quote} 

for the localizing approach, and for the labeling approach:
\begin{quote}
``How many \texttt{[category name]} are there in the image?''
\end{quote} 

\noindent
The prompt for the direct numerical estimation approach was instead defined by the following system message:
\begin{quote}
``You are attending a test on numerical perception. Perform your best numerical ability and respond to the question with just one number.''
\end{quote}

\noindent
and the user message:

\begin{quote}
``How many \texttt{[category name]} are there in the image?''
\end{quote}

\subsection{Prompts Explored}
We also explored different system prompts as well as user prompts to maximize the models' performance under each approach. We use separate validation sets for all datasets to decide which prompt results in the best performance.

\subsubsection{Label and count}
As for the label approach, we mainly focus on the system messages:

``You are attending a test for numerical perception and visual reasoning. You need to count how many objects there are in the image. Give each counted object a unique label and return the final count.''

``Please perform your best counting ability. Count from left to right, top to bottom. Assign each counted items a label and report the final count.''

\subsubsection{Point, label and count}
We also tried "Detect all the \texttt{[category name]} in the image." for user message and ''Detect the all of the prominent items in the image. The box\_2d should be [ymin, xmin, ymax, xmax] normalized to 0-1000.'' for system message.




\section{Threshold Optimization}\label{sec:threshold}
For PseCo, we used separate validation sets for all datasets by applying a grid search from 0.01 to 0.99 with an increment of 0.01 to find the best threshold on the metrics. The optimal thresholds for PseCo were determined as follows: for FSC-147, a threshold of 0.48 achieved an MAE of 3.7 and an Acc. of 0.28; for FSCD-LVIS, a threshold of 0.68 resulted in an MAE of 14.06 and an Acc. of 0.03; and for SolidCount, a threshold of 0.60 yielded an MAE of 6.28 and an Acc. of 0.09. These optimal thresholds were obtained using a separate, validation dataset, which is not always available in real-world scenarios. Thus, we report the performance of PseCo without those fine-tunings in the main text.



\section{Challenging Examples of FSCD-LVIS}\label{sec:fscd_challenge}
\begin{figure*}[t]
    \centering
    \includegraphics[width=0.9\linewidth, trim=20 10 20 10, clip]{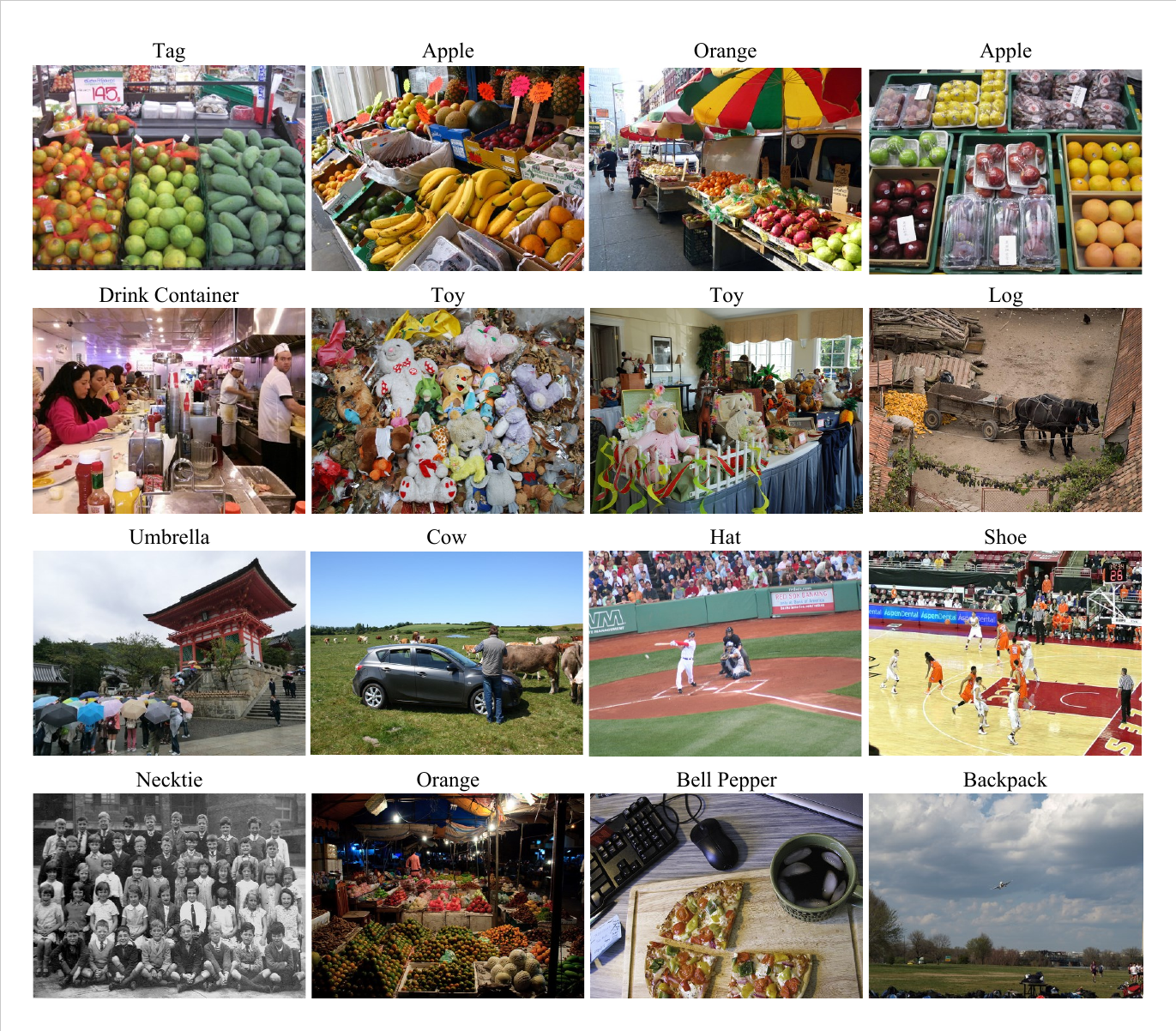}
    \caption{Examples of practically intractable counting cases from FSCD-LVIS dataset. On top of each image is the target object’s category name. These images exhibit extreme occlusion, low object visibility, and ambiguous object boundaries, making even human annotation unreliable and consistent counting nearly infeasible.}
    \label{fig:FSCD-LVIS Difficult Samples}
\end{figure*}

In this section, we discuss a set of practically intractable counting cases from the FSCD-LVIS dataset as shown in Fig. \ref{fig:FSCD-LVIS Difficult Samples}. These samples encompass various sources of complexity: (1) heavy occlusion and object clustering, such as in the Toy and Fruit categories, where objects are densely packed and partially overlap, making individual instances hard to distinguish; (2) low object salience, where targets are visually inconspicuous due to color similarity with the background (e.g., Drink Container) or low contrast conditions (e.g., Necktie in a grayscale group photo); (3) truncated or partially visible objects at image boundaries, which introduce ambiguity in determining the presence or completeness of targets (e.g., Cow, Backpack); and (4) ambiguous category definitions or uncertain countability, such as Log, where the boundaries of what constitutes a single instance are unclear. Due to the near-impossibility of obtaining consistent counts from these examples, we argue that such samples are unsuitable for reliable evaluation. We therefore recommend shifting focus toward the construction of more controllable and clearly defined datasets to facilitate meaningful progress in object counting research.

\begin{figure}
    \centering
    \includegraphics[width=0.99\linewidth]{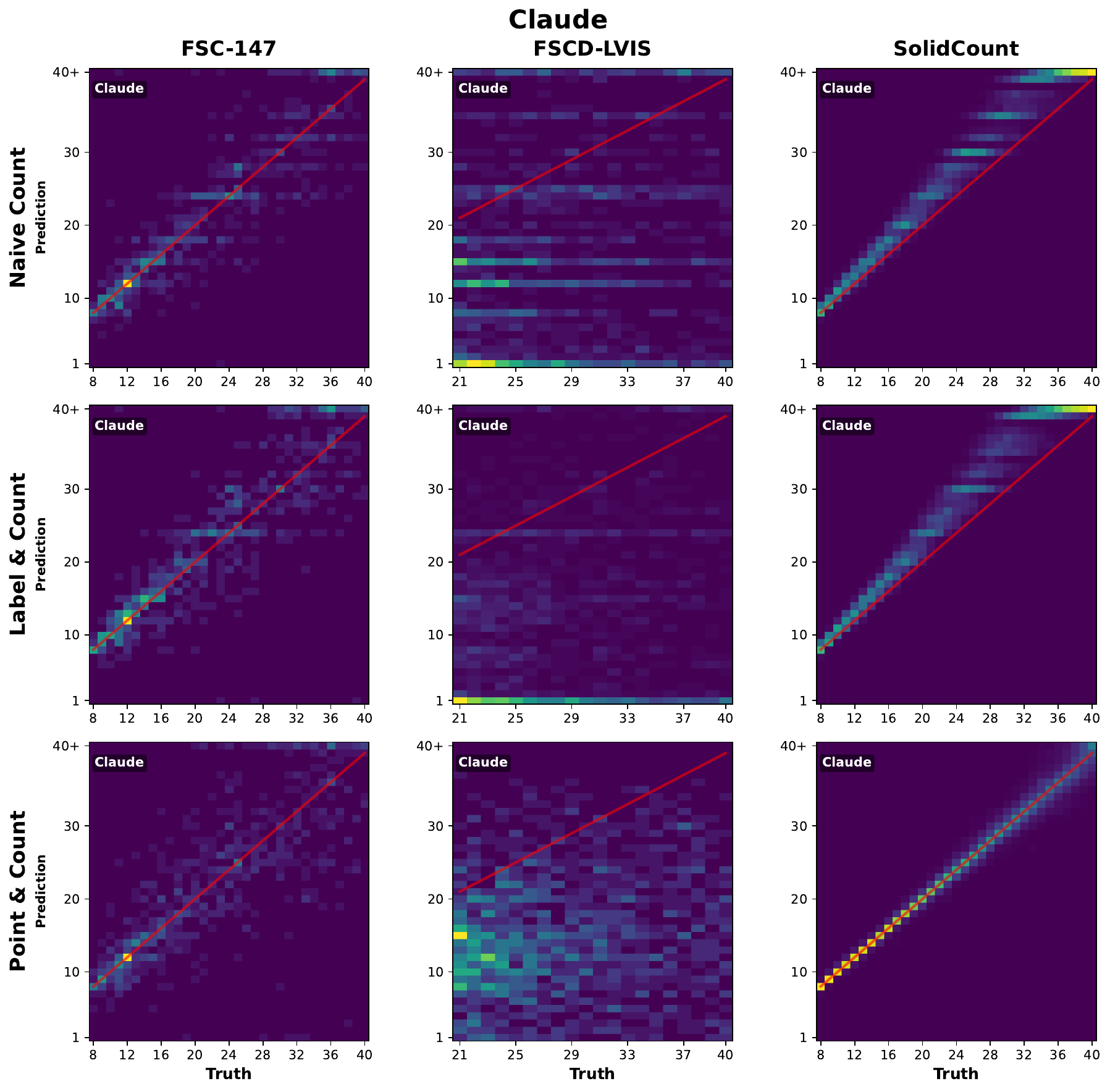}
    \caption{Confusion matrices across all datasets. The x-axis represents the ground truth, while the y-axis denotes the predictions, with the final row ("40+") aggregating all predictions where $n > 40$. The red line indicates the perfect responses.}
    \label{fig:placeholder}
\end{figure}
\begin{figure}
    \centering
    \includegraphics[width=0.99\linewidth]{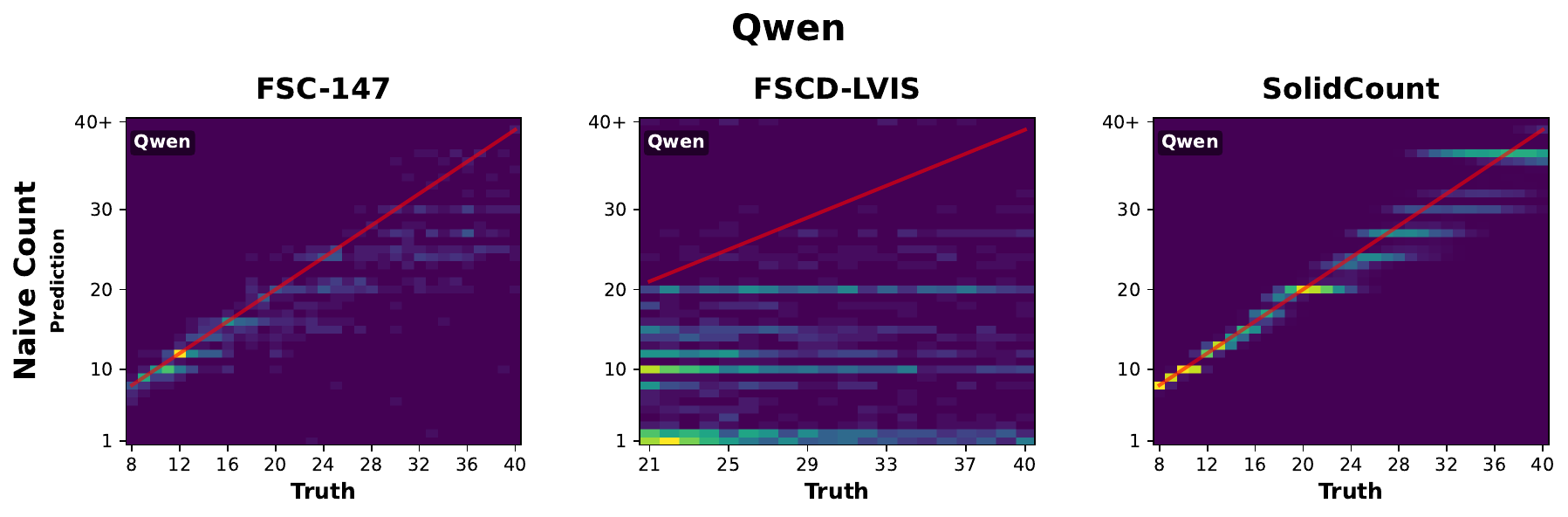}
    \caption{Confusion matrices across all datasets. The x-axis represents the ground truth, while the y-axis denotes the predictions, with the final row ("40+") aggregating all predictions where $n > 40$. The red line indicates the perfect responses.}
    \label{fig:placeholder}
\end{figure}
\begin{figure}
    \centering
    \includegraphics[width=0.99\linewidth]{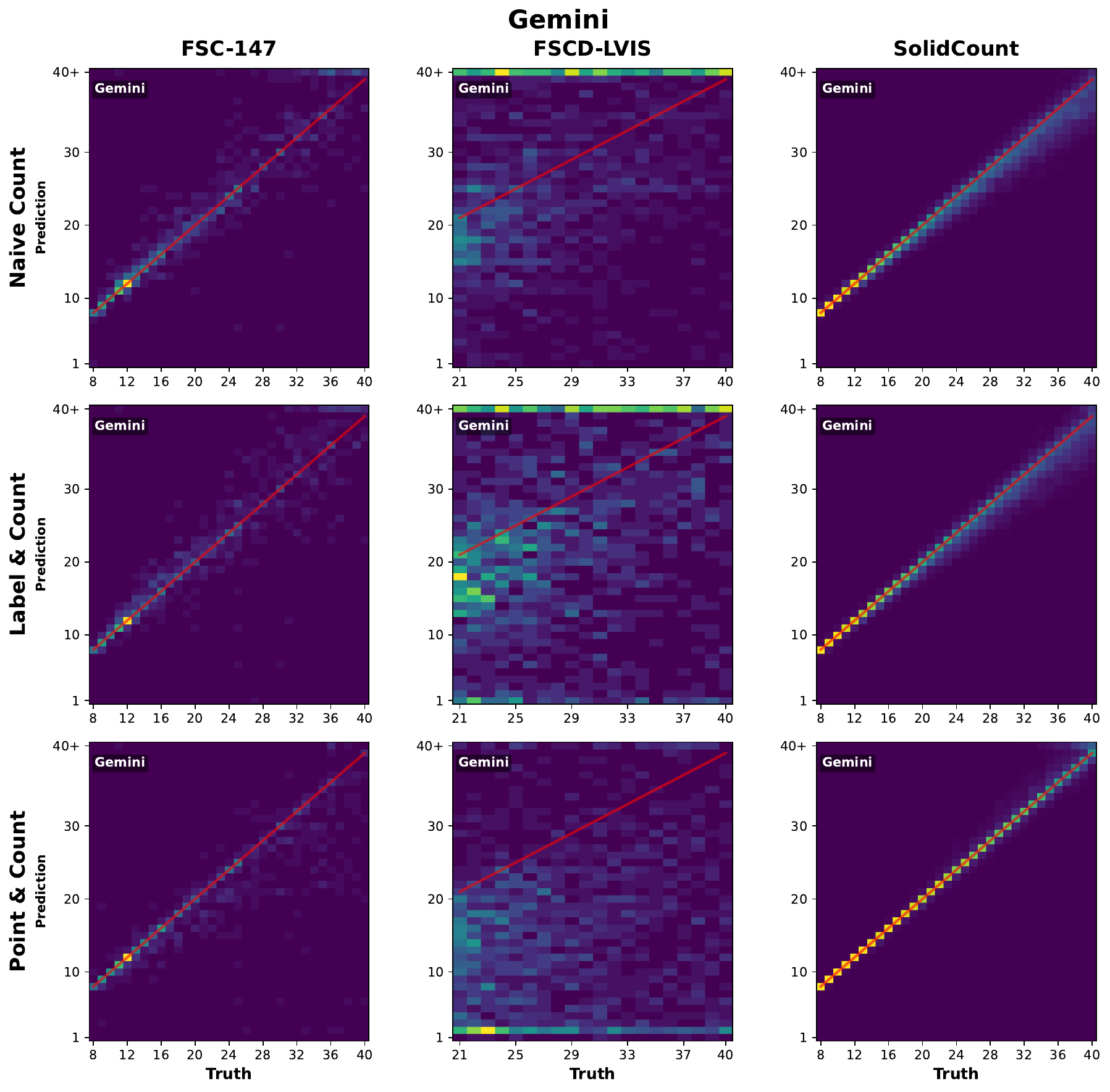}
    \caption{Confusion matrices across all datasets. The x-axis represents the ground truth, while the y-axis denotes the predictions, with the final row ("40+") aggregating all predictions where $n > 40$. The red line indicates the perfect responses.}
    \label{fig:placeholder}
\end{figure}
\begin{figure}
    \centering
    \includegraphics[width=0.99\linewidth]{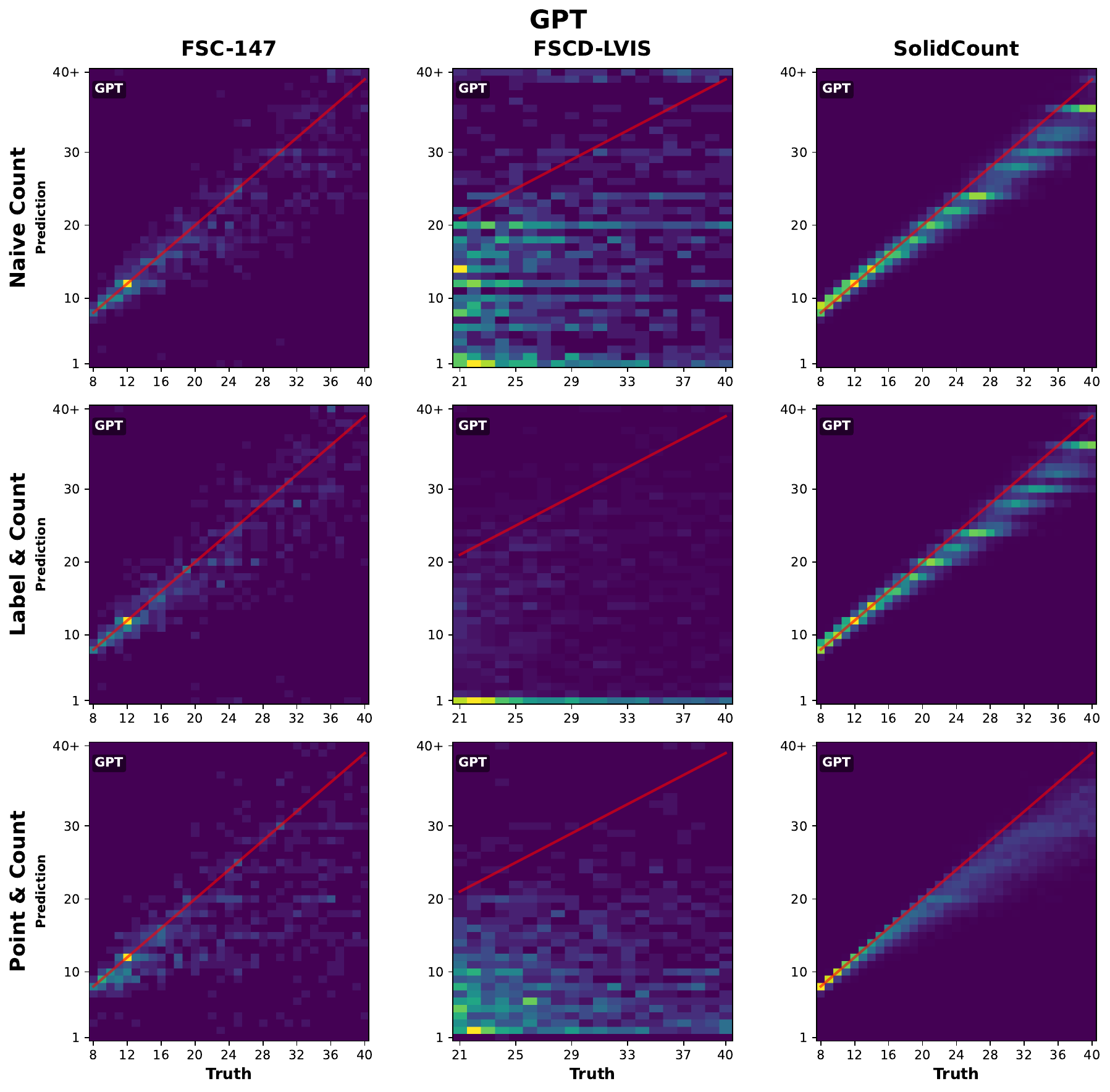}
    \caption{Confusion matrices across all datasets. The x-axis represents the ground truth, while the y-axis denotes the predictions, with the final row ("40+") aggregating all predictions where $n > 40$. The red line indicates the perfect responses.}
    \label{fig:placeholder}
\end{figure}
\begin{figure}
    \centering
    \includegraphics[width=0.99\linewidth]{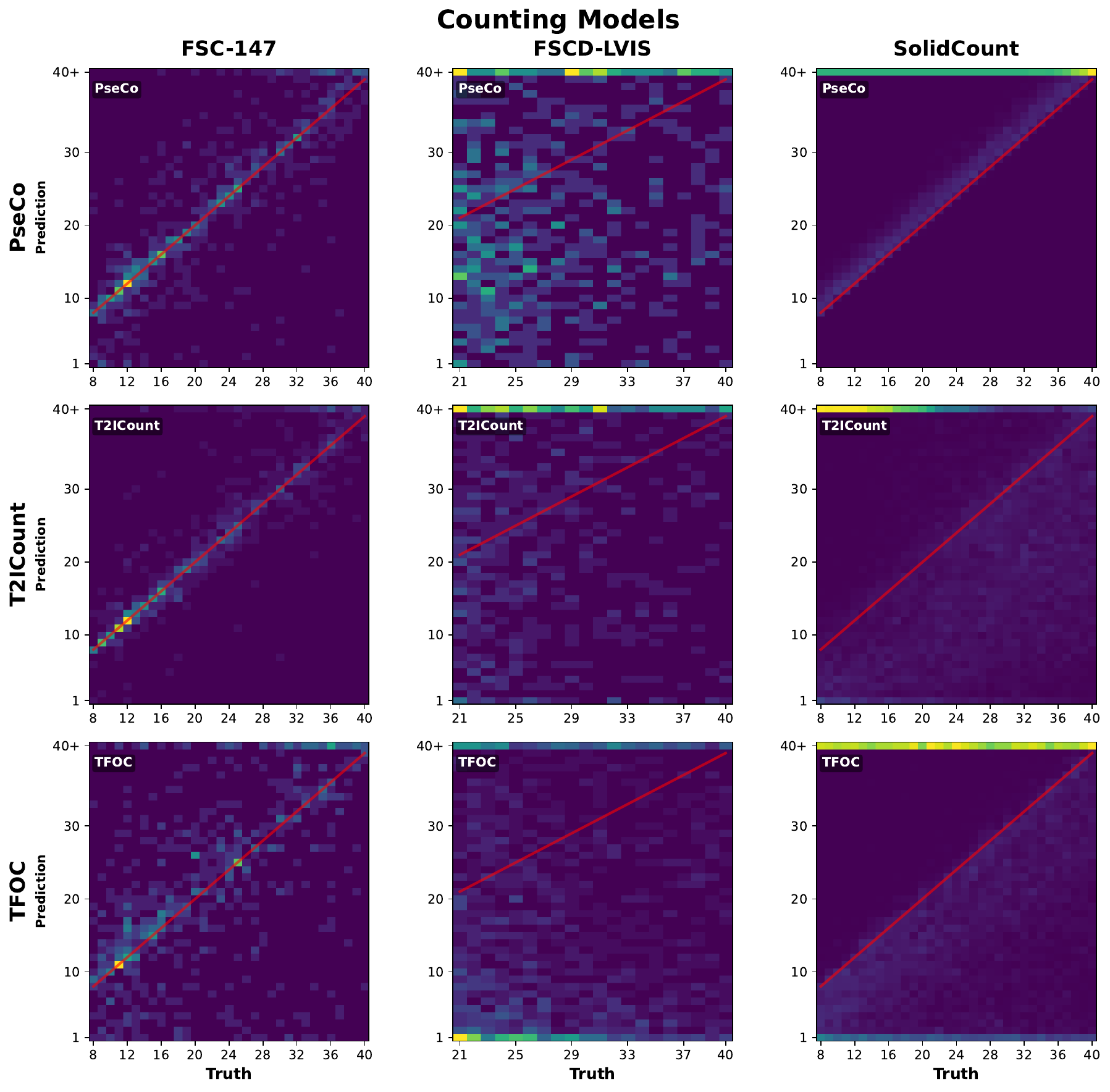}
    \caption{Confusion matrices across all datasets on \textbf{Counting Models}. The x-axis represents the ground truth, while the y-axis denotes the predictions, with the final row ("40+") aggregating all predictions where $n > 40$. The red line indicates the perfect responses.}
    \label{fig:placeholder}
\end{figure}




\begin{figure}
    \centering
    \includegraphics[width=0.99\linewidth]{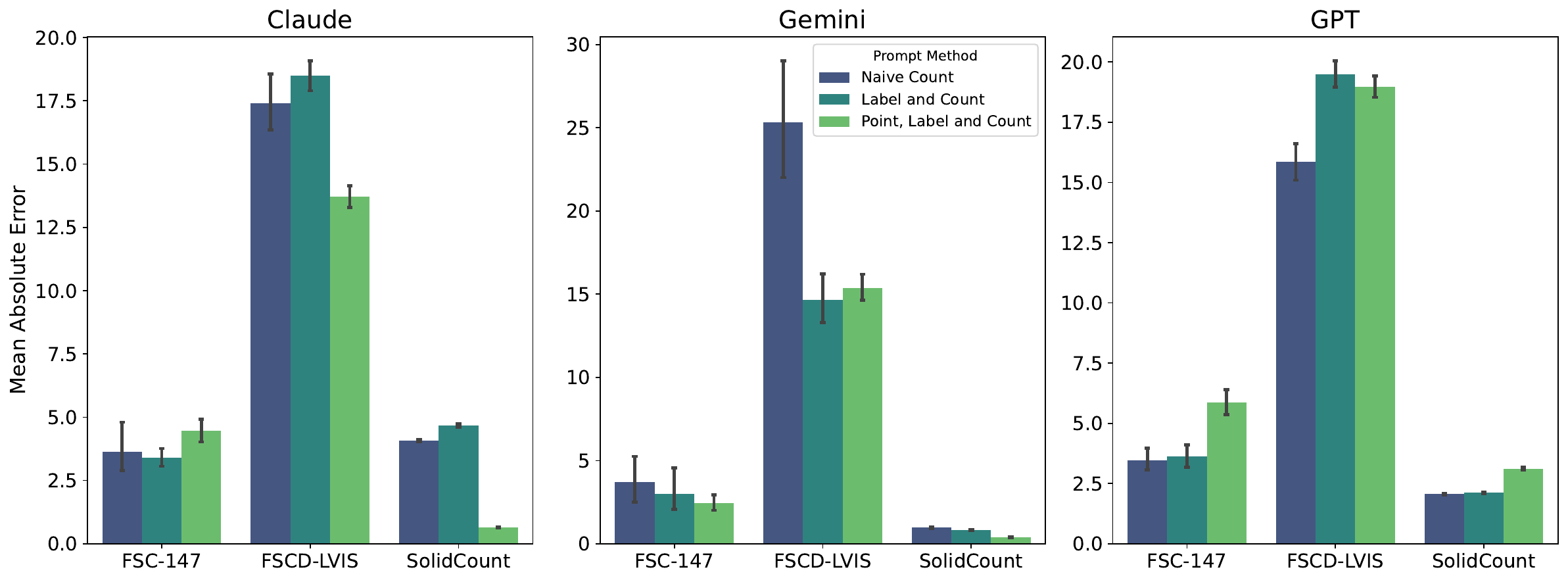}
    \caption{Bar chart displaying the MAE of three prompting strategies across all datasets. The height of each bar represents the mean MAE, and the error bars indicate the 95\% confidence interval of the mean.}
    \label{fig:mae_prompt_method}
\end{figure}

\end{document}